\pdfoutput=1
\documentclass[11pt]{article}

\usepackage{graphicx}
\usepackage{subcaption}

\usepackage{mathtools}
\graphicspath{ {./images} }

\usepackage[final]{acl}

\usepackage{rotating}
\usepackage{times}
\usepackage{latexsym}
\usepackage{booktabs}

\usepackage[T1]{fontenc}
\usepackage[utf8]{inputenc}

\usepackage{microtype}

\usepackage{inconsolata}

\title{Mitigating the Impact of Reference Quality on Evaluation of Summarization Systems with Reference-Free Metrics}

\author{Théo Gigant\footnotemark[1]\footnotemark[3], Camille Guinaudeau\footnotemark[2], Marc Decombas\footnotemark[3], Frederic Dufaux\footnotemark[1] \\
  \footnotemark[1] Université Paris-Saclay, CNRS, CentraleSupélec, Laboratoire des signaux et systèmes  \\
  \texttt{\{theo.gigant, frederic.dufaux\}@l2s.centralesupelec.fr}
  \\
  \footnotemark[2] Université Paris-Saclay, Japanese French Laboratory for Informatics, CNRS \\
  \texttt{guinaudeau@limsi.fr} \\
  \footnotemark[3] JustAI \\
  \texttt{marc@justai.co}\\
  }

\begin{document}
\maketitle
\begin{abstract}

% To evaluate automatic summaries, most frameworks ask human to provide fine-grained annotations to score the summaries on different qualities such as fluency, faithfulness, coherence, or relevance.
Automatic metrics are used as proxies to evaluate abstractive summarization systems when human annotations are too expensive. To be useful, these metrics should be fine-grained, show a high correlation with human annotations, and ideally be independent of reference quality; however, most standard evaluation metrics for summarization are reference-based, and existing reference-free metrics correlate poorly with relevance, especially on summaries of longer documents. In this paper, we introduce a reference-free metric that correlates well with human evaluated relevance, while being very cheap to compute. We show that this metric can also be used alongside reference-based metrics to improve their robustness in low quality reference settings.

\end{abstract}

\section{Introduction}

Given an input source, an abstractive summarization system should output a summary that is short, relevant, readable and consistent with the source.
To reflect this, fine-grained human evaluations are split into different scores \cite{fabbri_summeval_2021}, such as fluency, faithfulness (sometimes called factual consistency), coherence and relevance.
Fluency measures the linguistic quality of individual sentences, \textit{eg} if they contain no grammatical errors.
Coherence gauges if sentences in a summary are well-organized and well-structured. Faithfulness, or factual consistency, considers factual alignment between a summary and the source. Relevance is the measure of whether a summary contains the main ideas from the source.

Automatic summarization metrics are intended to capture one or multiple of these qualities \cite{zhu_gruen_2020, vasilyev_estime_2021}, and used as a proxy to evaluate summarization systems when human annotations are too expensive.

These metrics can be compared on their different attributes such as the reliance on one or multiple references, the cost of inference \cite{wu_less_2024}, the dataset-agnosticism \cite{faysse_revisiting_2023} and their correlations with human judgment at system-level \cite{deutsch_re-examining_2022} or summary-level.

In this work, we introduce a new reference-free metric \footnote{The code is available on \href{https://www.github.com/giganttheo/importance-based-relevance-score}{github}} that intends to capture the relevance of machine summaries using $n$-gram importance weighting.
We rate $n$-grams of the source documents relative to how much semantic meaning they express, as measured by \textit{tf-idf} \cite{sparck_jones_statistical_1972}, and score summaries according to their weighted lexical overlap with these $n$-grams.

We show that this metric is complementary to other metrics and can be mixed with reference-based metrics to alleviate their sensitivity to noisy and low quality references.

\section{Related Work}

\subsection{Extractive summarization using word-importance estimation}

A substantial amount of existing work investigated automatic extractive summarization using word-importance scores, based for instance on word statistics \cite{luhn_automatic_1958}, topic signatures \cite{lin_automated_2000} or pretrained models \cite{hong_improving_2014}.
Our approach follows a similar line of thought by utilizing a word-importance score to identify and weigh the $n$-grams that should be included in an abstractive summary with high relevance.

\subsection{Reference-based evaluation}

Lexical overlap based metrics such as ROUGE \cite{lin_rouge_2004}, BLEU \cite{papineni_bleu_2002} and chrF \cite{popovic_chrf_2015}, or pretrained language model based metrics such as BERTScore \cite{zhang_bertscore_2019} and BARTScore \cite{yuan_bartscore_2021}, are the standard way of evaluating abstractive summarization systems.
However, these metrics rely on gold standard reference summaries that can be costly, noisy, or missing altogether. We discuss some of the limits of these methods in section \ref{limits-of-ref}.

\subsection{LLM-as-a-Judge evaluation}

Large Language Models (LLMs) can perform many tasks effectively, even in few-shot or zero-shot settings. Recently, LLMs have also been used to evaluate natural language generation tasks, in replacement of human evaluation.
LLM-as-a-Judge shows useful properties as an evaluation metric, for instance \citet{faysse_revisiting_2023} illustrated using GPT-4 that it can be highly correlated with human judgement, format and task agnostic and comparable across tasks.
\citet{zheng_judging_2023} describe limitations of LLM-as-a-Judge, including position, verbosity and self-enhancement biases as well as poor performance at grading math or reasoning tasks. Other limitations are expressed by \citet{kim_prometheus_2023} targeting proprietary LLMs such as GPT-4 for their closed source nature, uncontrolled versioning, and their high costs. Prometheus 2 \cite{kim_prometheus_2024} is designed for evaluating language models and shows high correlations with proprietary LLMs and human evaluations. Besides, its open-source nature mitigates some of the aforementioned issues. \citet{liu_revisiting_2023} suggest that LLMs aligned from human feedback overfit to reference-less human evaluation of summaries, which they observed to be biased towards longer summaries and to suffer from low inter-annotator agreement.

\subsection{Reference-free evaluation}

Metrics designed to evaluate summaries without reference are useful when no gold reference are available, or when the property they intend to capture does not need a reference to be conveniently estimated.

GRUEN \cite{zhu_gruen_2020} aims at estimating the linguistic quality of a given summary by taking into account the grammaticality, non-redundancy, focus, structure and coherence of a summary.
% These attributes can be assessed without relying on a reference summary or even the source document. %GRUEN is using a weighted mix of $n$-gram based scores and multiple learnt scores.
ESTIME \cite{vasilyev_estime_2021} is evaluating the inconsistencies between the summary and the source by counting the mismatched embeddings out of the hidden layer of a pretrained language model. %, \textit{ie} the tokens of the summary that also appear in the text, and which closest embedding in the text are a different source token. 
% \citet{vasilyev_consistency_2022} introduce a generalized method based on ESTIME that is applicable to text-summary pairs with low extractive coverage and density. They also argue that this issue does not apply to current styles of summarization, which often have a high dictionary overlap.
Info Diff \cite{egan_play_2022} uses a pretrained model to compute the difference of Shannon information content between the source document and the source document given the summary.
 % is less focused on evaluating one specific aspect of summary quality, as it shows consistently high system-level correlations with expert-annotated coherence, consistency, fluency and relevance scores. It 
FEQA \cite{durmus_feqa_2020} and SummaQA \cite{scialom_answers_2019} both compare how a model answers to questions about the document given the source document or a proposed summary.

% According to \citet{liu_revisiting_2023}, human evaluations often suffer from low inter-annotator agreement, especially crowd-workers among themselves and with expert annotators.
% Their Atomic Content Units protocol gives a better inter-annotator agreement. In their RoSE benchmark, they provide human annotations on different settings, using only the summary, using the source document, a reference, and the Atomic Content Units.
\citet{liu_revisiting_2023} observed that reference-free human evaluations have a very low correlation with reference-based human evaluations, and tend to be biased towards different types of systems. %They showed that with reference-free evaluation the few-shot model GPT-3 scored highest while reference-based evaluation ranked the smaller abstractive summarization models higher.

\subsection{Evaluating Summarization of Long Documents}

Trained metrics usually generalize poorly to out-of-distribution tasks \cite{koh_how_2022}, and often cannot handle long contexts. In the long document summarization setting, \citet{koh_how_2022} showed that most automatic metrics correlate poorly with human judged relevance and factual consistency scores. \citet{wu_less_2024} use an extract-then-evaluate method to reduce the size of the long source document used as a reference for evaluation of factual consistency and relevance with LLM-as-a-Judge. They find that it both lowers the cost of evaluation, and improve the correlation with human judgement.

\section{Limits of reference-based evaluation}
\label{limits-of-ref}

Lexical overlap scores such as BLEU or ROUGE work under the implicit assumption that reference summaries are mostly extractive and contain no errors.
This assumption is challenged by a study conducted by \citet{maynez_faithfulness_2020} on hallucinated content in abstractive summaries. In human written summaries from the XSum dataset, $76.9\%$ of the gold references were found to have at least one hallucinated word.

Summarization methods can trade abstractiveness for faithfulness, creating a faithfulness-abstractiveness tradeoff curve that was illustrated and studied by \citet{ladhak_faithful_2022}. They show that some metrics are more sensitive to the summary abstractiveness than others. %, along a selector approach to improve this tradeoff, resulting in higher \textit{effective faithfulness}. As shown in the paper, some metrics are more sensitive to the summary abstractiveness than others.

In the context of translations, \textit{translationese} refers to source language artifacts found in both human and machine translations. This phenomenon is similar to extractive segments in summaries, as it is an artifact of the source document that can be mitigated through paraphrasing. \citet{freitag_bleu_2020} demonstrated that reference translations in machine translation datasets tend to exhibit this \textit{translationese} language. They addressed this by creating new references through paraphrasing the existing ones. When tested, systems produced much lower BLEU scores with the paraphrased references compared to the \textit{translationese} ones, but the correlation with human judgment was higher. They observed that with \textit{translationese} references, the $n$-grams with the highest match rates resulted from translations adhering to the source sentence structure. In contrast, using the paraphrased references, the most-matched $n$-grams were related to the semantic meaning of the sentence.

Following a \textit{translationese} - extractiveness analogy, we assume that with highly extractive references, the most matched $n$-grams between proposed and reference summaries are artifacts of the extractiveness of the summaries. More abstractive references will yield much lower ROUGE scores, but might correlate better with human judgement.

We propose to use $n$-gram importance weighting methods, such as \textit{tf-idf} \cite{sparck_jones_statistical_1972} or \textit{bm-25} \cite{robertson_relevance_1976}, to extract the $n$-grams expressing most of the semantic meaning of the source document. We believe that these $n$-grams should appear in relevant summaries, and are not artifacts of extractiveness.

\section{Proposed Metric}

\label{sec:proposed-metric}

% We design our metric such that:
% \begin{itemize}
%     \item It is cheap to compute
%     \item It has a good 
%     \item A useful range
%     \item 
% \end{itemize}

% Let $f_{t,d}$ be the frequency of a $n$-gram $t$ in a document $d$ from a corpus $D$.
Let $W_{t,d,D}$ be the importance of a $n$-gram $t$ in a document $d$ from a corpus $D$, defined as
\[
    W_{t,d,D} = 
\begin{dcases}
     \tanh({\frac{w_{t,d,D}}{r_{t,d,D}}}),& \text{if } t\in d \\
     0, &\text{otherwise,}
\end{dcases}
\]

$w_{t,d,D}$ is an importance score obtained through word importance scoring methods (such as \textit{tf-idf} and \textit{bm-25}). The associated importance rank of the $n$-gram in the document is referred as $r_{t,d,D}$.

Given a proposed summary $\hat{s}$ of a document $d \in D$, we compute the metric:
$$m(\hat{s}, d, D) = \frac{\alpha_{\hat{s}, d, D}}{N_{d, D}} \Sigma_{t \in \hat{s}} W_{t,d,D}$$ 

With $N_{d, D}$ the upper ceiling of the sum of weights, used to normalize the score: $N_{d, D}~=~\Sigma_{t \in d}~W_{t,d,D}$.

% By construction this score will be maximum for a summary consisting of the full document. To alleviate this issue, we penalize longer summaries by multiplying with a function $f$ accounting for the length of the summary $|\hat{s}|$ and the length of the document $|d|$: $\alpha_{\hat{s}, d} = f({|\hat{s}|},{|d|})$\footnote{The choice for $f$ is illustrated in Appendix \ref{appendix:range-values}, Figure \ref{fig:lengthpenalty}}.

By design this score will be maximized for a summary consisting of the full document. To alleviate this issue, we penalize longer summaries by multiplying with a term accounting for the length of the summary $|\hat{s}|$ relative to the length of the document $|d|$: $\alpha_{\hat{s}, d} = f({|\hat{s}|},{|d|})$\footnote{The choice for $f$ is illustrated in Appendix \ref{appendix:range-values}, Figure \ref{fig:lengthpenalty}}.
We observe that this length penalty not only resolves the issue related to the scoring of entire documents but also shows a stronger correlation with human judgment at the system level.

It is relatively straightforward to devise a trivial heuristic that achieves a high score by employing the same $n$-gram importance weighting method to generate an extractive summary, with access to the full corpus.
We do not consider this point to be a substantial issue, as such heuristic will result in a low score on metrics that measure other aspects of an abstractive summary, such as fluency.

\section{Experiments}

For our experiments, we work with different datasets of human evaluation of summarization systems.
SummEval \cite{fabbri_summeval_2021} contains human evaluations for $23$ systems, each with $100$ summaries of news article from the CNN/DailyMail dataset. Coherence, consistency, fluency and relevance are evaluated by experts and crowd-source workers. ArXiv and GovReport \cite{koh_how_2022} contain annotations for $12$ summarization systems, evaluated on $18$ long documents for each dataset. Human evaluators rated the factual consistency and the relevance of the machine summaries. RoSE \cite{liu_revisiting_2023} is a benchmark consisting of $12$ summarization systems evaluated on $100$ news article from CNN/DailyMail. Each summary is annotated with different protocols, we are using the reference-based and reference-free human evaluations.

We describe the choice of settings for our metric in Appendix \ref{appendix:range-values}, which takes into account system-level correlations on the four datasets, as well as the range of values taken by the metric.

\subsection{System-level correlation scaling with number of summaries}

% According to \citet{deutsch_re-examining_2022}, system-level correlations are usually inconsistent with the practical use of automatic evaluation metrics. When computing these correlations to evaluate systems, usually only the subset of summaries judged by humans is used. However automatic metrics can be computed on summaries outside of this subset to give better estimates.  
% With our reference-free metric we could go one step further and evaluate the systems on more documents, even without reference summaries.
% \citet{deutsch_re-examining_2022} also illustrates that testing with more examples will narrow down the confidence intervals of the evaluated scores, making it more convenient to compare systems.

\begin{figure}[h]
    \centering
    \begin{subfigure}{.7\linewidth}
        \centering
        \includegraphics[width=.9\linewidth]{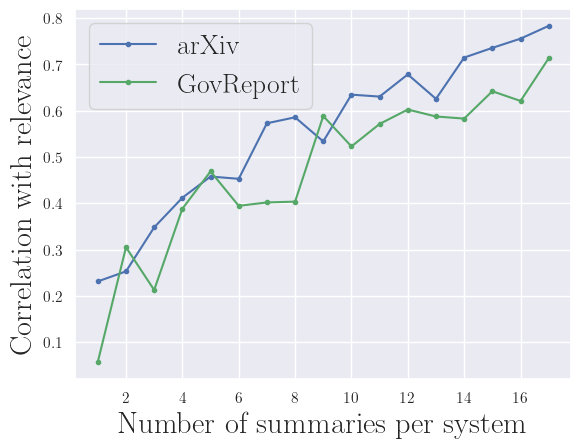}
        \caption{ArXiv and GovReport} \label{fig:arxivgovreport-scaling}
    \end{subfigure}
    % \hspace*{\fill}
    \begin{subfigure}{.7\linewidth}
        \centering
        \includegraphics[width=.9\linewidth]{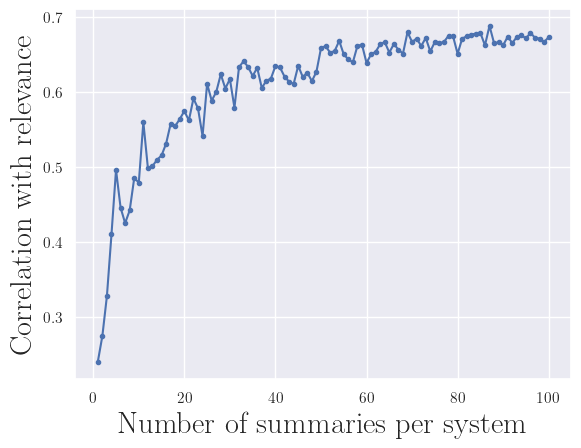}
        \caption{SummEval} \label{fig:summeval-scaling}
    \end{subfigure}
    \caption{System-level correlations with human judgement for our metric, depending on the number of summaries used for evaluation}
    \label{fig:corr-scaling}
\end{figure}

According to \citet{deutsch_re-examining_2022}, system-level correlations are usually inconsistent with the practical use of automatic evaluation metrics. To evaluate systems, usually only the subset of summaries judged by humans is used. However automatic metrics can be computed on summaries outside of this subset to give better estimates. \citet{deutsch_re-examining_2022} also illustrates that testing with more examples will narrow down the confidence intervals of the evaluated scores, making it more convenient to compare systems.
With a reference-free metric like ours, systems can be evaluated on more documents without the need for reference summaries.
Figure \ref{fig:corr-scaling} illustrates the increase of system-level correlation with human evaluated relevance when using more examples for each system. 

% Figure \ref{fig:corr-scaling} shows an increase of the system-level correlation with human evaluated relevance when using more examples for each system. 

% \begin{figure}[h]
%     \begin{subfigure}{0.48\linewidth}
%         \centering
%         \includegraphics[width=.9\linewidth]{scaling-correlations-arxivgovreport.png}
%         \caption{ArXiv and GovReport} \label{fig:arxivgovreport-scaling}
%     \end{subfigure}
%     \hspace*{\fill}
%     \begin{subfigure}{0.48\linewidth}
%         \centering
%         \includegraphics[width=.9\linewidth]{scaling-correlations-summeval.png}
%         \caption{SummEval} \label{fig:summeval-scaling}
%     \end{subfigure}
%     \caption{System-level correlations with human judgement for our metric, depending on the number of summaries used for evaluation}
%     \label{fig:corr-scaling}
% \end{figure}

\subsection{Robustness to noisy references}

\begin{figure}[h]
    \centering
    % \begin{subfigure}{0.9\linewidth}
    %     \centering
    %     \includegraphics[width=.9\linewidth]{rouge-altered-refs-summeval-rand.png}
    % \caption{SummEval}
    % \end{subfigure}
    % \hfill
    \begin{subfigure}{\linewidth}
        \centering
        \includegraphics[width=.9\linewidth]{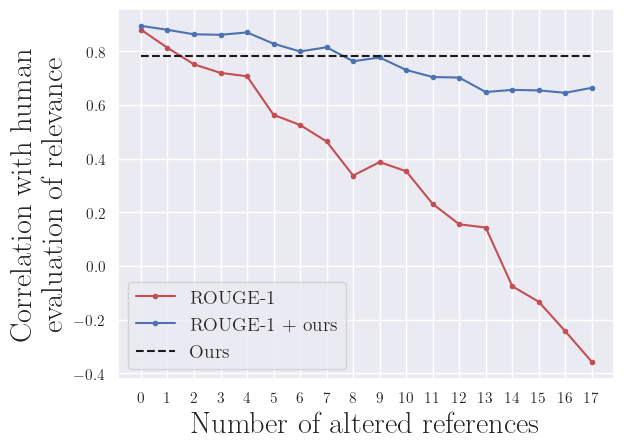}
        % \caption{arXiv}
    \end{subfigure}
    % \begin{subfigure}{0.8\linewidth}
    %     \centering
    %     \includegraphics[width=.9\linewidth]{rouge-altered-refs-govreport.png}
    %     \caption{GovReport}
    % \end{subfigure}
    \caption{System-level correlation with human evaluation of relevance, depending on the number of altered references (RAND-3 alteration).}
    \label{fig:altered-rouge}
\end{figure}

Reference-based metrics such as ROUGE-1 are sensitive to the quality of the references. To evaluate the robustness of ROUGE-1 to noisy references, we gradually replace random reference summaries with altered references and compute the resulting system-level correlations. The references are altered by replacing them with three random sentences (RAND-3) from the source document. Results with the ArXiv dataset, averaged over 20 random draws, are reported in Figure \ref{fig:altered-rouge}. Results with different alteration methods and different datasets are reported in Figures \ref{fig:altered-rouge-rand}, \ref{fig:altered-rouge-lead} and \ref{fig:altered-rouge-tail} in Appendix \ref{appendix:range-values}.
Our metric is not sensitive to altered references by design, contrary to ROUGE-1. When mixed with it, it improves the robustness of ROUGE-1 to low quality references. This aspect is beneficial in settings where the quality of the reference summaries is unknown or variable, for instance with web-crawled datasets.

% We observe a surprising behaviour of ROUGE-1 where its system-level correlations are improved on average when introduced to alterations in the range of $0$ to $30\%$ of the samples of the SummEval dataset. This behaviour does not seem to translate to the mix of ROUGE-1 with our metric.

% \begin{figure}[h]
%     \centering    \includegraphics[width=.7\linewidth]{rouge-altered-refs.png}
%     \caption{System-level correlation depending on the number of altered references in SummEval}
%     \label{fig:altered-rouge}
% \end{figure}

\subsection{Complementarity with other automatic metrics}

\begin{figure}[h]
    \centering
    \includegraphics[width=.8\linewidth]{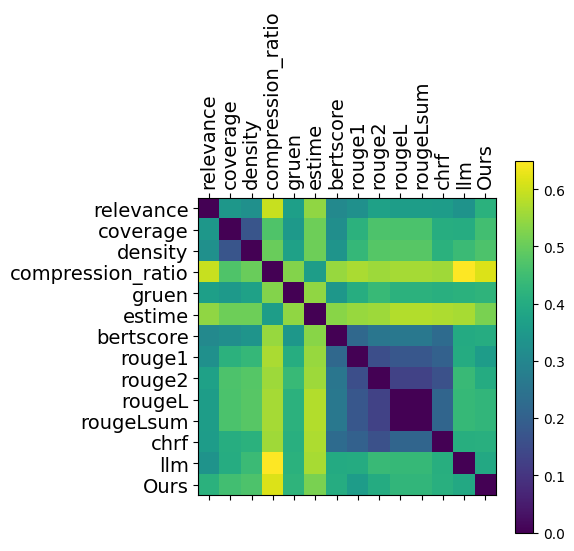}
    \caption{Complementarity between metrics on SummEval}
    \label{fig:complementarity}
\end{figure}

We report the pairwise complementarity between each pair of metric\footnote{we use the \href{https://github.com/huggingface/evaluate}{\texttt{evaluate}} implementation of ROUGE, chrF and BERTScore and official implementations of \href{https://github.com/WanzhengZhu/GRUEN}{GRUEN} and \href{https://github.com/PrimerAI/blanc}{ESTIME}} on SummEval in Figure \ref{fig:complementarity}, following \citet{colombo_glass_2023}. We observe that our metric has a high complementarity with most other metrics, noticeably with ROUGE and chrF scores, which are also based on lexical overlap, meaning that they capture different features of the evaluated summaries.

\begin{table}[h]
\centering
\caption{System-level correlations of mixes of metrics}
\begin{tabular}{l|c|c|c}
\label{tab:mixes}
     \small Metric &  \small SummEval & \small ArXiv & \small GovReport  \\
     \hline
     \small ROUGE-1 &  0.59 & \textbf{0.88} & \textbf{0.92} \\
     \small ROUGE-2 & 0.61 & 0.52 & \textbf{0.91} \\
     \small ROUGE-L & 0.47 & 0.72 & 0.90 \\
     \small chrF & \textbf{0.75} & 0.83 & 0.87 \\
     \small BERTScore & 0.40 & 0.32 & \textbf{0.91} \\
     \small{ROUGE-1 + chrF} & \textbf{0.75} & \textbf{0.89} & 0.90
      \\
     \hline
     \small ESTIME & -0.45 & 0.18 & -0.69 \\
     \small GRUEN & 0.59 & 0.32 & -0.37 \\
     \small LLM-as-a-judge & \textbf{0.88} & \textbf{0.76} & 0.63 \\
     \small \textbf{Ours} & 0.67 & \textbf{0.78} & \textbf{0.71} \\
     \hline
     \small{\textbf{Ours} + ROUGE-1} & \textbf{0.80} & \textbf{0.90} & \textbf{0.85}
      \\
     \small{\textbf{Ours} + chrF} & 0.74 & 0.83 & 0.82
      \\
     \small{\textbf{Ours} + BERTScore} & 0.74 & 0.77 & 0.76
      \\
     \hline
     \small{LLM + ROUGE-1} & \textbf{0.91} & \textbf{0.90} & \textbf{0.85}
      \\
     \small{LLM + chrF} & 0.89 & 0.87 & \textbf{0.85}
      \\
     \small{LLM + BERTScore} & \textbf{0.91} & 0.81 & 0.70
      \\
     \hline
     \small{\textbf{Ours} - ESTIME} & {0.71} & { -0.01} & {\textbf{0.77}}
      \\
     \small{\textbf{Ours} + GRUEN} & \textbf{0.83} & \textbf{0.71} & -0.12
      \\
     \hline
\end{tabular}
\end{table}

In Table \ref{tab:mixes} we report the system-level Spearman correlations using our metric, other metrics, and simple combinations of metrics.
In the LLM-as-a-judge method, we are using the \texttt{gemini-1.5-flash} model \cite{gemini_team_gemini_2024} following the prompt proposed by \citet{wu_less_2024} to evaluate the relevance of summaries.

Our simple metric achieves comparable results to LLM-as-a-Judge methods in term of correlations with human evaluations of summary relevance across various settings, at a significantly lower cost.

\section{Conclusion and future works}

In this work, we introduce a new reference-free metric based on importance-weighted $n$-gram overlap between the summary and the source. We demonstrated that it has high correlations with human judgement and can be used alongside other metrics to improve them and mitigate their sensitivity to low-quality references.

The prospects for future research include further exploration of the behaviour of reference-based, reference-free and hybrid metrics with references of varying quality, as well as potential extensions to multimodal settings such as the evaluation of vision-language systems.

\section{Limitations}

Like other lexical overlap metrics, ours works with the assumption that there is a vocabulary overlap between the source document and the summary, \textit{ie} that the summary has a non-zero coverage. In order to evaluate the sensitivity of our metric to various levels of extractiveness of summaries, we would have wanted to compute the score on systems with varying values on the faithfulness-abstractiveness tradeoff curve presented in \citet{ladhak_faithful_2022}; but their data was not made available yet.

\citet{vasilyev_is_2021} noticed that higher correlation with human scores can be achieved with "false" improvements, mimicking human behaviour. Using a referenceless evaluation metric, they limited the comparisons with the source text by selecting sentences to maximize their score, and observed a higher correlation with human judgement as a result. \citet{wu_less_2024} observe a similar consequence by first extracting sentences that maximize the ROUGE score with the original document and using the resulting extracted sentences along with the predicted summary as the input to be evaluated by a LLM-as-a-judge. Their interpretation however is different as they do not view this higher correlation with human scores as a "false" improvement, but as a way to mitigate the \textit{Lost-in-the-Middle} problem of LLMs.

We believe that the relevant interpretation depends on the method that is used to extract sentences from the source document.
Using comparisons with the summary to extract "oracle" spans of the original document, or selecting key sentences that span over the main information of the document are not motivated by the same reasons. Mimicking the human behaviour of referring only to the bits of the document that are relevant to the proposed summary \textit{at first glance} to score marginally higher correlations is a different thing than filtering the most important bits of a document relative to a measure of word importance.

Our metric filters out the $n$-grams with little semantic significance in the document. This can mimick the human bias of comparing the summary to salient sentences only, but it will also lower the influence of the artifacts of extractiveness discussed in section \ref{limits-of-ref}.

Our metric is also specific to the task of summarization and might correlate differently with human judgement on summarization tasks with different compression ratio, extractiveness, or style. Table \ref{tab:spur} in the Appendix \ref{appendix:range-values} illustrates this.

LLM-as-a-Judge methods can solve the issues of sensitivity to extractiveness and task settings, while providing more interpretable results, but are not exempt from biases and come with a noticeably higher cost.

\bibliography{custom}

% \newpage
\clearpage

\appendix

\section{Appendix}

\label{appendix:range-values}

\subsection{Spurious correlations}

\citet{durmus_spurious_2022} observed that model-based reference-free evaluation often has higher correlations with spurious correlates such as perplexity, length, coverage or density, than with human scores. We report the correlations between metrics and spurious correlates in Table \ref{tab:spur}.

\subsection{Correlations with human judgement on different settings}

\begin{figure*}[h]
\centering
    \begin{subfigure}{0.3\linewidth}
        \includegraphics[width=\linewidth]{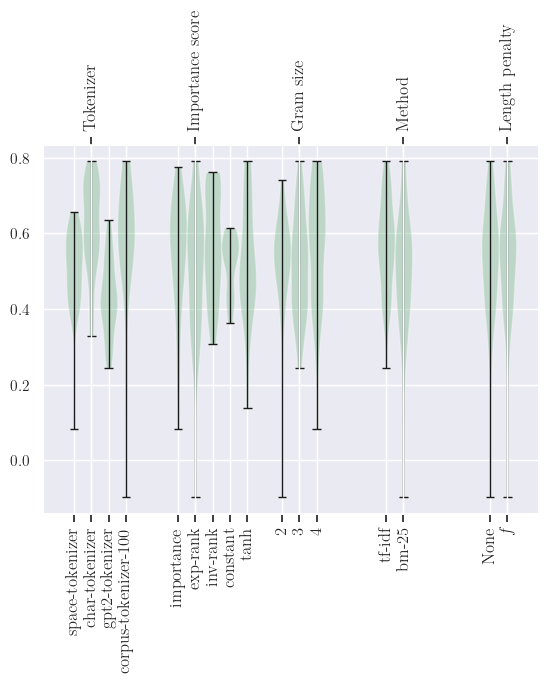}
        \caption{ArXiv}
    \end{subfigure}
    % \hspace*{\fill}
    \begin{subfigure}{0.3\linewidth}
        \includegraphics[width=\linewidth]{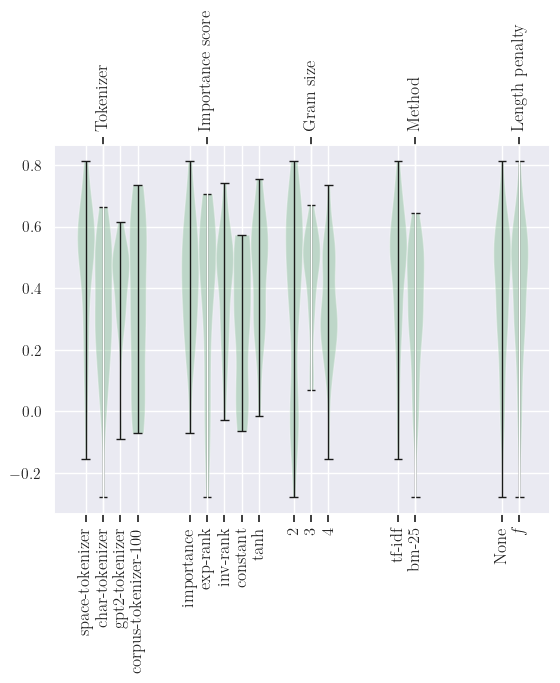}
        \caption{GovReport}
    \end{subfigure}

    \begin{subfigure}{0.3\linewidth}
        \includegraphics[width=\linewidth]{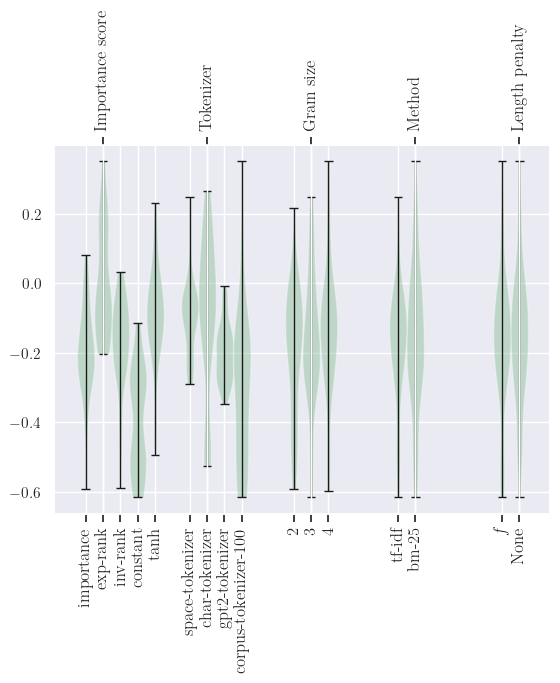}
        \caption{RoSE (reference-based)}
    \end{subfigure}
    % \hspace*{\fill}
    \begin{subfigure}{0.3\linewidth}
        \includegraphics[width=\linewidth]{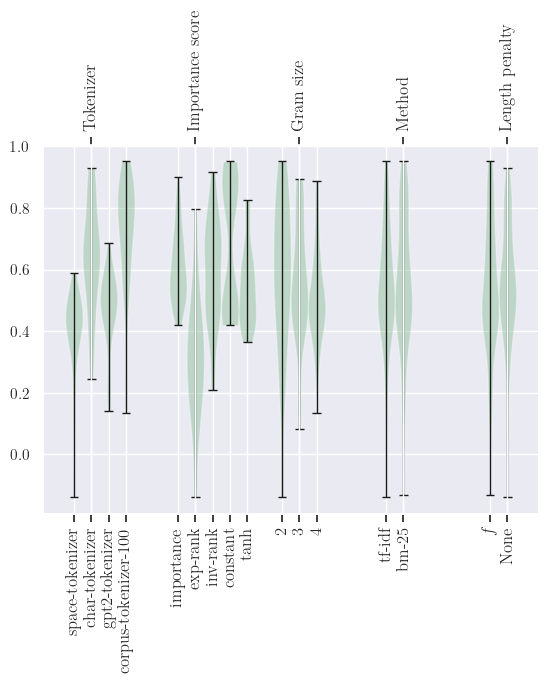}
        \caption{RoSE (reference-free)}
    \end{subfigure}

    \begin{subfigure}{0.3\linewidth}
        \includegraphics[width=\linewidth]{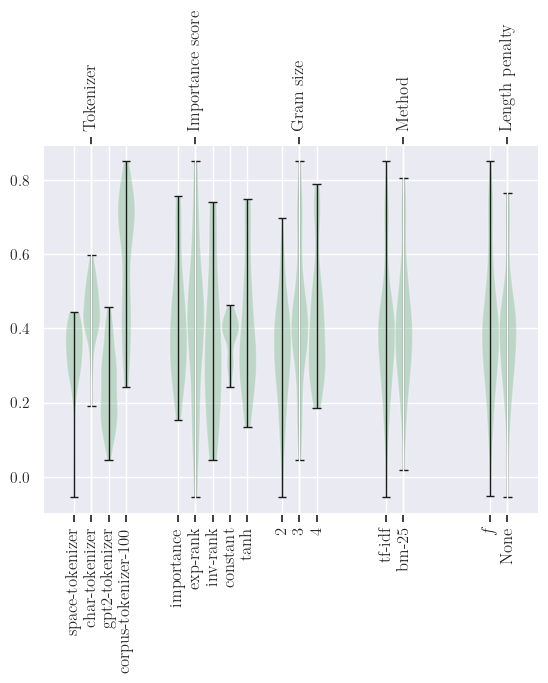}
        \caption{SummEval}
    \end{subfigure}
    \newline

    \caption{ Distribution of system-level correlations of our metric in different settings}
\label{fig:distr}
\end{figure*}

Figure \ref{fig:distr} illustrate the distributions of system-level correlations of our metric with different settings.

For tokenization, we tested tokenizing texts as separated by space, using character tokenization, a pretrained GPT-2 tokenizer, or a custom tokenizer, trained on each corpus with a vocabulary of $100$ tokens.

We included different sizes of $n$-grams in our tests, with bigrams, trigrams and $4$-grams.

The two methods we considered for importance weigthing are \textit{tf-idf} and \textit{bm-25}.

The importance score is the weight used to score the overlapped $n$-grams, we included the following scores:
\begin{itemize}
    \item importance: $t,d,D \mapsto w_{t,d,D}$
    \item exp-rank: $t,d,D \mapsto \exp(-r_{t,d,D})$
    \item inv-rank: $t,d,D \mapsto \frac{1}{r_{t,d,D}}$
    \item constant: $t,d,D \mapsto 1$
    \item tanh: $t,d,D \mapsto \tanh(\frac{w_{t,d,D}}{r_{t,d,D}})$
\end{itemize}

\begin{figure}[h]
    \includegraphics[width=\linewidth]{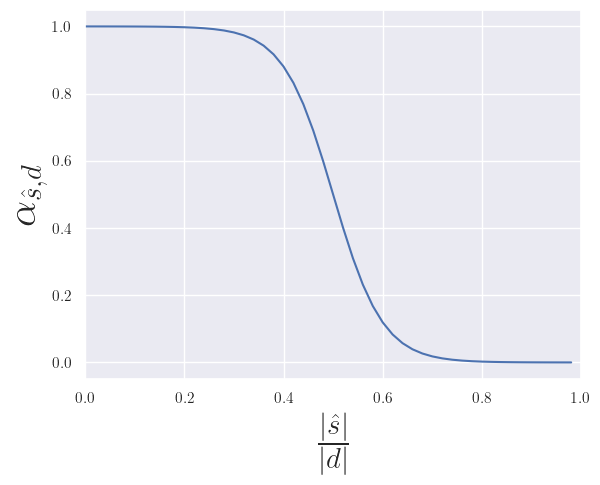}
    \caption{
    Length penalty $ \alpha_{ \hat{s}, d} = f({| \hat{s}|},{|d|}) $ with \newline
    $ f: | \hat{s}|, | \hat{d}| \mapsto \frac{1}{1 + \exp(20 * \frac{| \hat{s}|}{|d|} - 10) } $
    }
    \label{fig:lengthpenalty}
\end{figure}

The options for the length penalty $\alpha_{\hat{s}, \hat{d}}$ are no penalty or  $\alpha_{\hat{s}, d} = f({|\hat{s}|},{|d|})$, with $$f: |\hat{s}|, |d| \mapsto \frac{1}{1 + \exp(20 * \frac{|\hat{s}|}{|d|} - 10)}$$
$f$ is illustrated in Figure \ref{fig:lengthpenalty}.

We chose to use the corpus tokenizer, with trigrams, \textit{tf-idf} and the tanh importance scoring with length penalty. These settings proved to be consistant in the tested conditions, and provided good ranges of values on different inputs. All the other experiments with our metric in this paper are using these settings.

\hspace*{\fill}

\begin{figure}[h]
    \centering
    \begin{subfigure}{\linewidth}
        \centering
        \includegraphics[width=.9\linewidth]{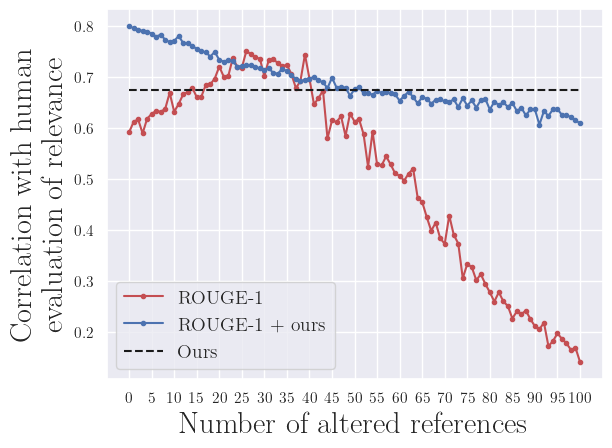}
    \caption{SummEval}
    \end{subfigure}
    \begin{subfigure}{\linewidth}
        \centering
        \includegraphics[width=.9\linewidth]{rouge-altered-refs-arxiv-rand.png}
        \caption{arXiv}
    \end{subfigure}
    \begin{subfigure}{\linewidth}
        \centering
        \includegraphics[width=.9\linewidth]{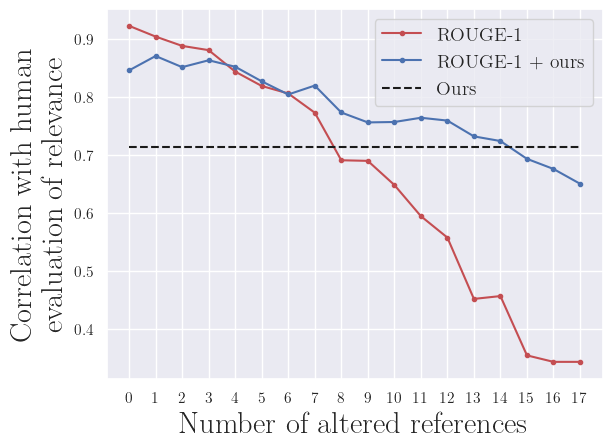}
        \caption{GovReport}
    \end{subfigure}
    \caption{System-level correlation with human evaluation of relevance, depending on the number of altered references (RAND-3 alteration).}
    \label{fig:altered-rouge-rand}
\end{figure}

\begin{figure}[h]
    \centering
    \begin{subfigure}{\linewidth}
        \centering
        \includegraphics[width=.9\linewidth]{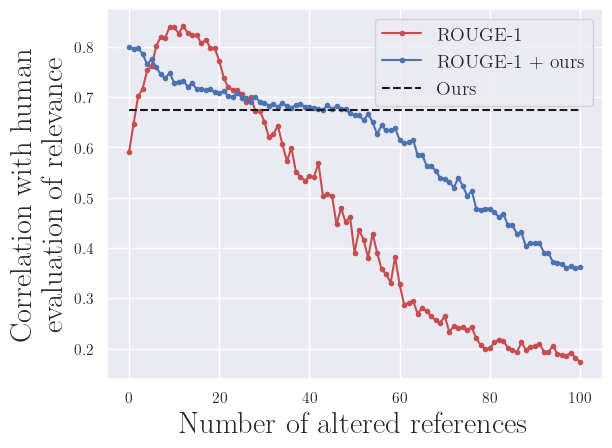}
    \caption{SummEval}
    \end{subfigure}
    \begin{subfigure}{\linewidth}
        \centering
        \includegraphics[width=.9\linewidth]{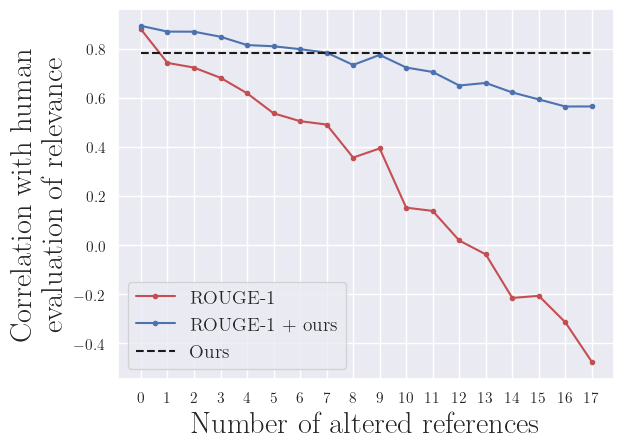}
        \caption{arXiv}
    \end{subfigure}
    \begin{subfigure}{\linewidth}
        \centering
        \includegraphics[width=.9\linewidth]{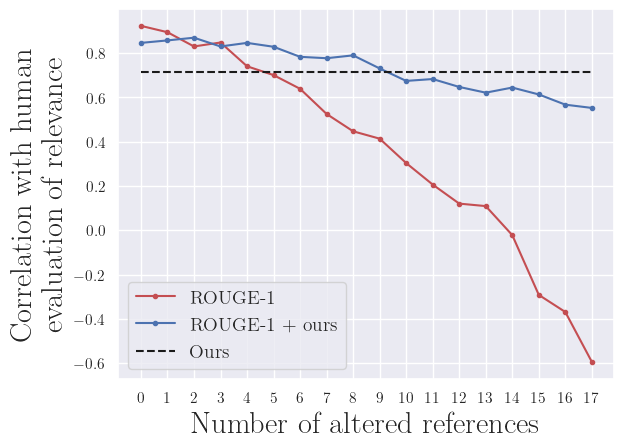}
        \caption{GovReport}
    \end{subfigure}
    \caption{System-level correlation with human evaluation of relevance, depending on the number of altered references (LEAD-3 alteration).}
    \label{fig:altered-rouge-lead}
\end{figure}

\begin{figure}[h]
    \centering
    \begin{subfigure}{\linewidth}
        \centering
        \includegraphics[width=.9\linewidth]{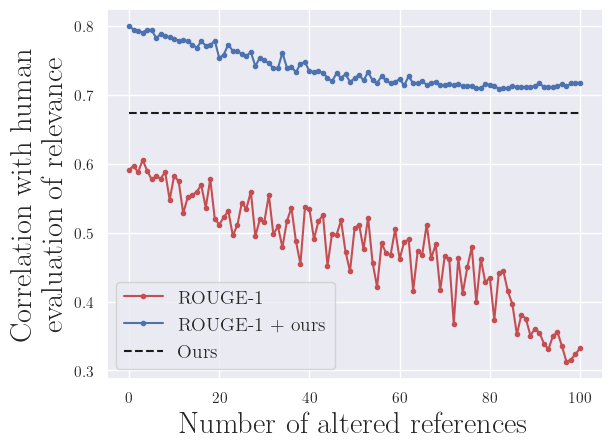}
    \caption{SummEval}
    \end{subfigure}
    \begin{subfigure}{\linewidth}
        \centering
        \includegraphics[width=.9\linewidth]{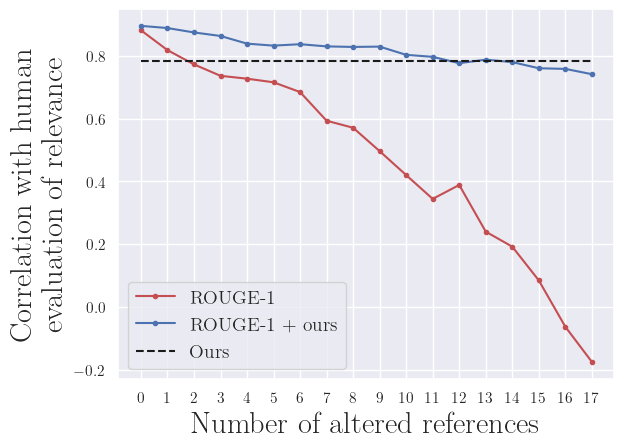}
        \caption{arXiv}
    \end{subfigure}
    \begin{subfigure}{\linewidth}
        \centering
        \includegraphics[width=.9\linewidth]{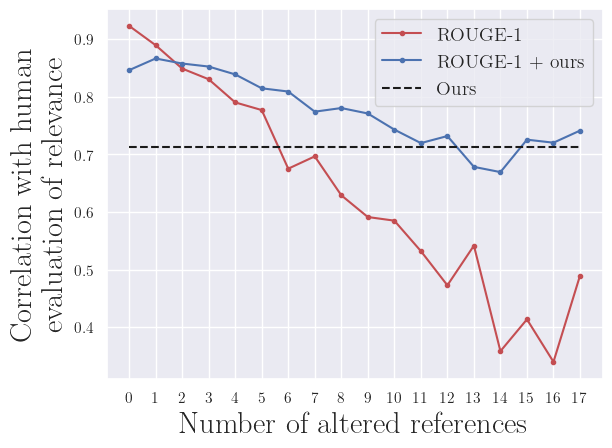}
        \caption{GovReport}
    \end{subfigure}
    \caption{System-level correlation with human evaluation of relevance, depending on the number of altered references (TAIL-3 alteration).}
    \label{fig:altered-rouge-tail}
\end{figure}

Figures \ref{fig:altered-rouge-rand}, \ref{fig:altered-rouge-lead} and \ref{fig:altered-rouge-tail} show the system-level correlation of our metric, ROUGE-1 and their combination as we gradually replace the reference summaries with respectively three random sentences (RAND-3), the first three (LEAD-3) or last three (TAIL-3) sentences of the source document.

\subsection{Range of values}

\begin{figure*}[h]
    \begin{subfigure}{0.48\linewidth}
        \includegraphics[width=\linewidth]{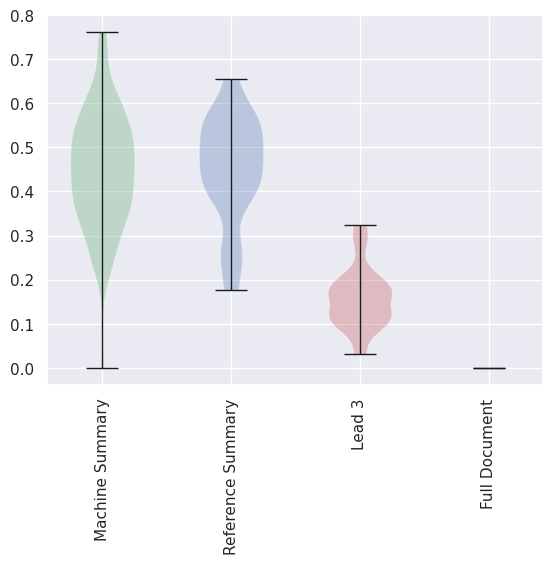}
        \caption{ArXiv}
    \end{subfigure}
    \hspace*{\fill}
    \begin{subfigure}{0.48\linewidth}
        \includegraphics[width=\linewidth]{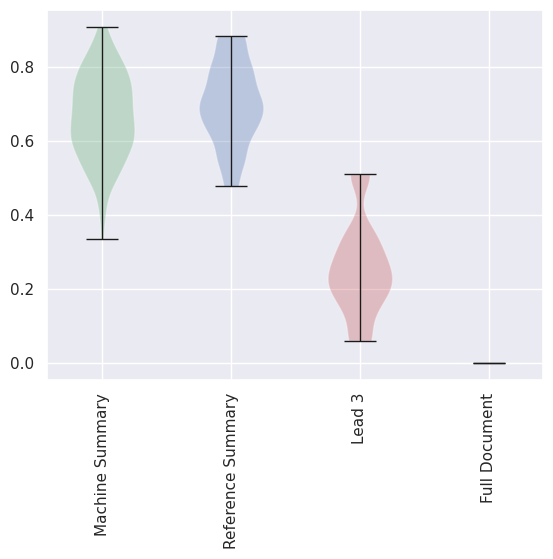}
        \caption{GovReport}
    \end{subfigure}
    
    \begin{subfigure}{0.48\linewidth}
        \includegraphics[width=\linewidth]{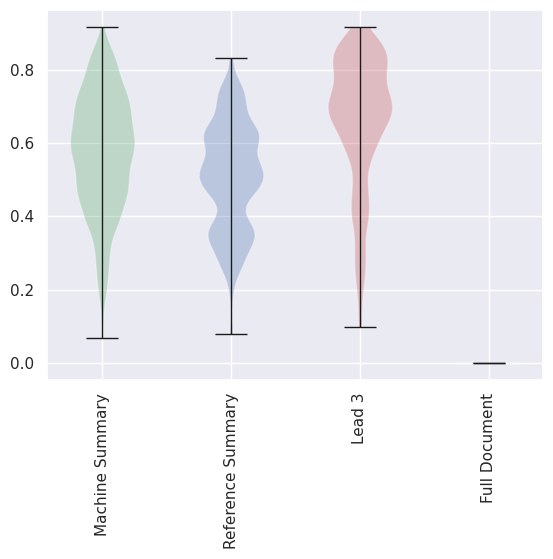}
        \caption{SummEval}
    \end{subfigure}
    \hspace*{\fill}
    \begin{subfigure}{0.48\linewidth}
        \includegraphics[width=\linewidth]{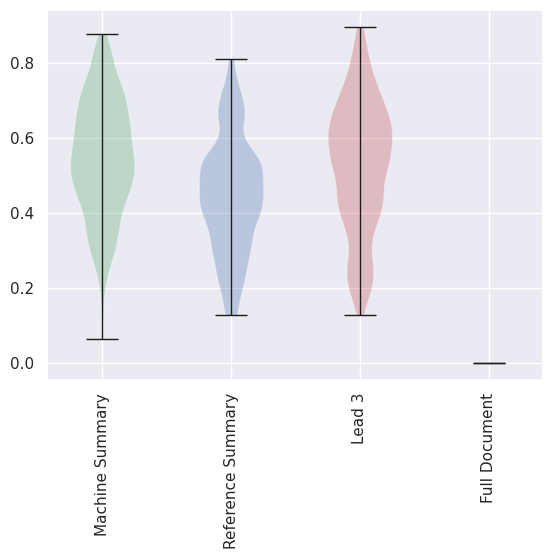}
        \caption{RoSE}
    \end{subfigure}
    \caption{Range of values taken by our metric for different summaries}
    \label{fig:values-ours}
\end{figure*}

\begin{figure*}[h]
    \begin{subfigure}{0.48\linewidth}
        \includegraphics[width=\linewidth]{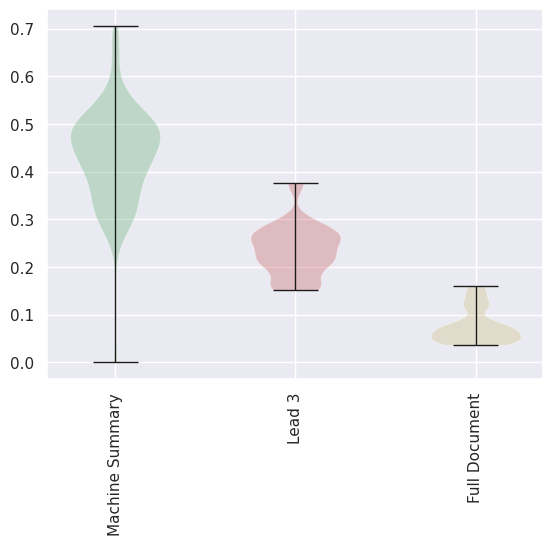}
        \caption{ArXiv}
    \end{subfigure}
    \hspace*{\fill}
    \begin{subfigure}{0.48\linewidth}
        \includegraphics[width=\linewidth]{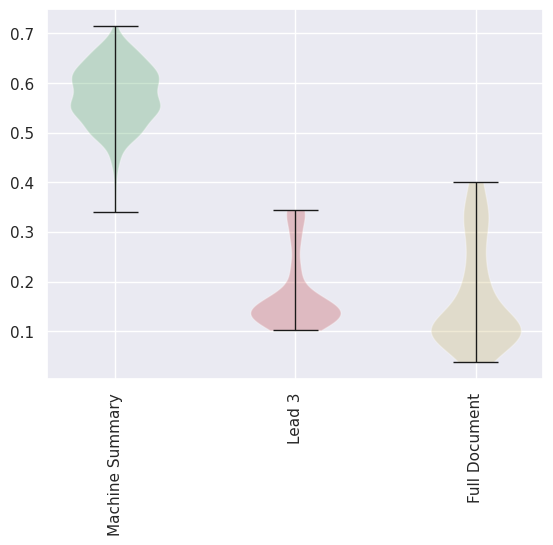}
        \caption{GovReport}
    \end{subfigure}
    
    \begin{subfigure}{0.48\linewidth}
        \includegraphics[width=\linewidth]{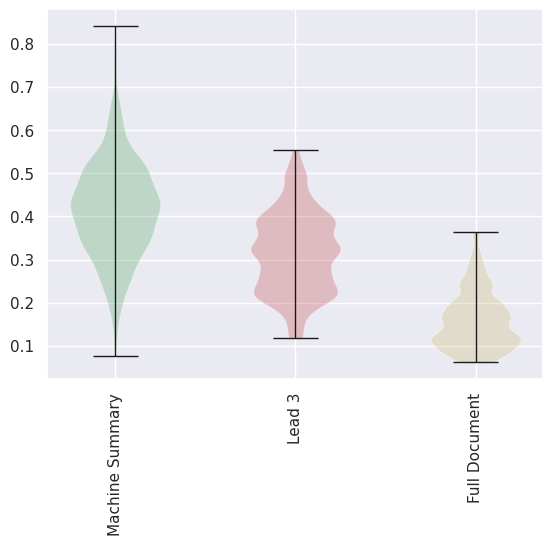}
        \caption{SummEval}
    \end{subfigure}
    \hspace*{\fill}
    \begin{subfigure}{0.48\linewidth}
        \includegraphics[width=\linewidth]{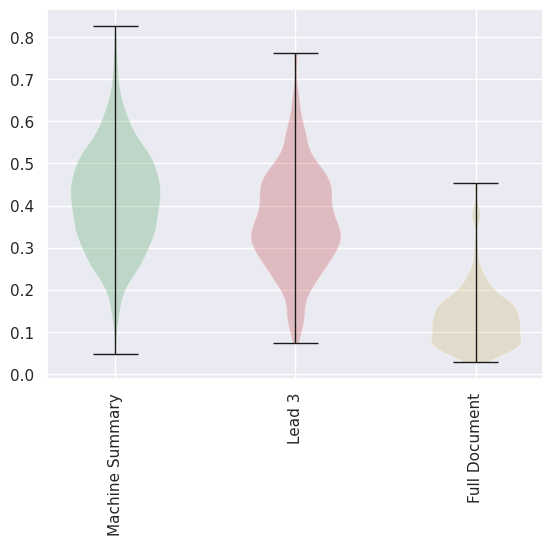}
        \caption{RoSE}
    \end{subfigure}
    \caption{Range of values taken by ROUGE-1 for different summaries}
    \label{fig:values-rouge}
\end{figure*}

We report the range of values taken by our metric, and ROUGE-1, for different inputs and on different datasets in Figures \ref{fig:values-ours} and \ref{fig:values-rouge}.

\begin{table*}
\caption{Summary-level correlations between our metric, human evaluation metrics and spurious correlates. Values are bolded when the correlation with spurious correlate is higher than with human evaluation.}
\begin{tabular}{l|c|c|c|c|c|c|c|c}
\label{tab:spur}
      &  \multicolumn{2}{c|}{SummEval} &  \multicolumn{2}{c|}{arXiv} &  \multicolumn{2}{c|}{GovReport} &  \multicolumn{2}{c}{RoSE} \\
     \hline
     Metric & \small Pearson & \small Spearman & \small Pearson & \small Spearman &  \small Pearson & \small Spearman & \small Pearson & \small Spearman  \\
     \hline
    Relevance & 0.24 & 0.23 & 0.42 & 0.45 & 0.15 & 0.18 & & \\
    Coherence & 0.20 & 0.23 & & & & & & \\
    Consistency & 0.03 & 0.04 & -0.02 & -0.11 & -0.30 & -0.32 & & \\
    Fluency & 0.01 & -0.01 & & & & & & \\
    Reference-based & & & & & & & 0.17 & 0.14 \\
    Reference-free & & & & & & & 0.18 & 0.15 \\
    \hline
    Coverage & \textbf{0.26} & \textbf{0.28} & 0.07 & 0.43 & \textbf{-0.20} & \textbf{-0.19} & -0.08 & 0.05 \\
    Density & 0.18 & 0.22 & -0.04 & -0.02 & -0.03 & 0.17 & 0.04 & 0.02 \\
    Compression Ratio & -0.03 & -0.06 & 0.01 & 0.02 & \textbf{-0.54} & \textbf{-0.64} & \textbf{-0.28} & \textbf{-0.18} \\
    Summary Length & \textbf{0.32} & \textbf{0.28} & 0.40 & 0.28 & 0.02 & -0.08 & \textbf{0.30} & \textbf{0.26} \\
     
\end{tabular}
\end{table*}

% ...

\end{document}